\documentclass{article}

\usepackage{microtype}
\usepackage{graphicx}
\usepackage{subfigure}
\usepackage{booktabs} % for professional tables
\usepackage{flushend}
\usepackage{caption}
\usepackage{chngcntr}
\usepackage{dblfloatfix}
\usepackage{datetime}

\usepackage{hyperref}

\usepackage[accepted]{icml2018}

\newcommand{\version}{}

\icmltitlerunning{Essentially No Barriers in Neural Network Energy Landscape\version}

\usepackage{amsmath}
\usepackage{amssymb}
\usepackage{amsthm}
\usepackage{bm}
\usepackage[noabbrev,capitalize]{cleveref}
\usepackage[T1]{fontenc}
\usepackage[utf8]{inputenc}

\usepackage[titletoc,title]{appendix}

\DeclareMathOperator*{\argmin}{arg\,min}

\let\oldvec\vec
\renewcommand{\vec}[1]{\oldvec{\mathit{#1}}}
\newcommand{\norm}[1]{\left\lVert#1\right\rVert}
\newcommand{\PLH}{{\mkern-2mu\times\mkern-2mu}}

\begin{document}

\twocolumn[
\icmltitle{Essentially No Barriers in Neural Network Energy Landscape\version}




\begin{icmlauthorlist}
\icmlauthor{Felix Draxler}{ial,thphys}
\icmlauthor{Kambis Veschgini}{thphys}
\icmlauthor{Manfred Salmhofer}{thphys}
\icmlauthor{Fred A. Hamprecht}{ial}
\end{icmlauthorlist}

\icmlaffiliation{ial}{Heidelberg Collaboratory for Image Processing (HCI), IWR, Heidelberg University, D-69120 Heidelberg, Germany}
\icmlaffiliation{thphys}{Institut für Theoretische Physik, Heidelberg University, D-69120 Heidelberg, Germany}

\icmlcorrespondingauthor{Fred A. Hamprecht}{fred.hamprecht@iwr.uni-heidelberg.de}

\icmlkeywords{Machine Learning, ICML}

\vskip 0.3in
]



\printAffiliationsAndNotice{}  % leave blank if no need to mention equal contribution

\begin{abstract}
	Training neural networks involves finding minima of a high-dimensional non-convex loss function.
	Relaxing from linear interpolations, we construct continuous paths between minima of recent neural network architectures on CIFAR10 and CIFAR100.
	Surprisingly, the paths are essentially flat in both the training and test landscapes.
	This implies that minima are perhaps best seen as points on a single connected manifold of low loss, rather than as the bottoms of distinct valleys.
\end{abstract}
\section{Introduction}
\label{sec:intro}

Neural networks have achieved remarkable success in practical applications such as object recognition~\cite{he16deep,huang17densely}, machine translation~\cite{bahdanau15neural,vinyals15neural}, speech recognition~\cite{hinton12deep,graves13speech,xiong17toward} etc.
Theoretical insights on why neural networks can be trained successfully despite their high-dimensional and non-convex loss functions are few or based on strong assumptions such as the eigenvalues of the Hessian at critical points being random~\cite{dauphin14identifying}, linear activations~\cite{choromanska14loss,kawaguchi16deep} or wide hidden layers~\cite{soudry16bad,nguyen17loss}.

In the current literature, minima of the loss function are typically depicted as points at the bottom of a strictly convex valley of a certain width that reflects the generalisation of the network, with network parameters given by the location of the minimum~\cite{keskar16large}.
This is also the picture obtained when the loss function of neural networks is visualised in low dimension~\cite{li17visualizing}.

%This is especially present in work about the generalisation gap, where it is claimed that wide minima generalise better~\cite{keskar16large}.

In this work, we conjecture that neural network loss minima are not isolated points in parameter space, but essentially form a connected manifold.
More precisely, we argue that the part of the parameter space where the loss remains below a certain low threshold forms one single connected component.

\begin{figure}[th]
	\centering
	\includegraphics[width=\linewidth]{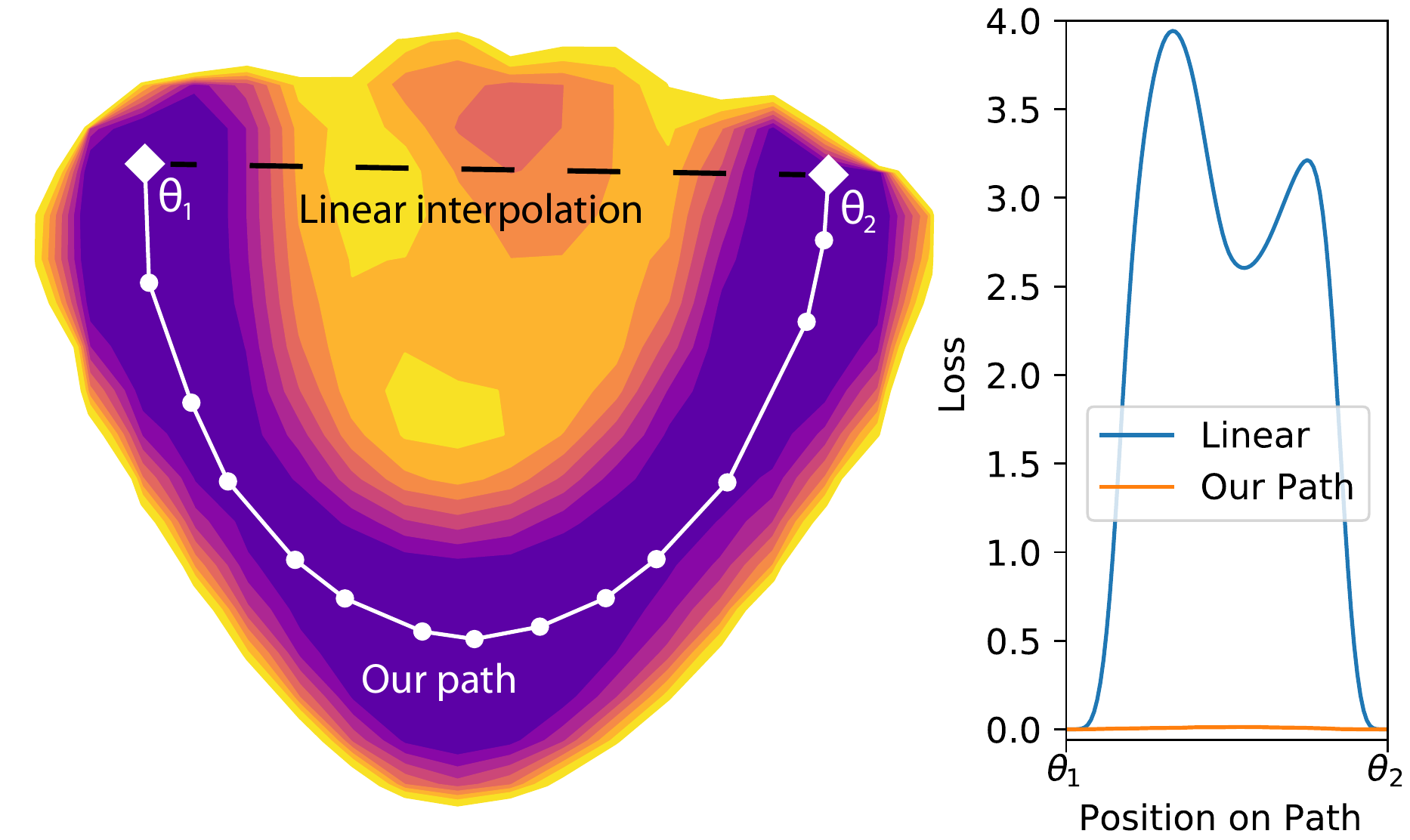}
	\vspace{-.5cm}
	\caption{\textit{Left:} A slice through the one million-dimensional training loss function of DenseNet-40-12 on CIFAR10 and the minimum energy path found by our method. The plane is spanned by the two minima and the mean of the nodes of the path. \textit{Right:} Loss along the linear line segment between minima, and along our high-dimensional path. Surprisingly, the energy along this path is essentially flat.}
	\label{fig:min_ene_path}
\end{figure}

We support the above claim by studying the energy landscape of several ResNets and DenseNets on CIFAR10 and CIFAR100:
For random pairs of minima, we construct continuous paths through parameter space for which the loss remains very close to the value found directly at the minima.
An example for such a path is shown in \cref{fig:min_ene_path}.

Our main contribution is the finding of paths
\begin{enumerate}
	\item that connect minima trained from different initialisations which are not related to each other via known loss-conserving operations like rescaling,
	\item along which the training loss remains essentially at the same value as at the minima,
	\item along which the test loss remains essentially constant while the test error rate slightly increases.
\end{enumerate}
The existence of such paths suggests that modern neural networks have enough parameters such that they can achieve good predictions while a big part of the network undergoes structural changes.
In closing, we offer qualitative justification of this behaviour that may offer a handle for future theoretical investigation.

%This work is structured as follows:
%In \cref{sec:related}, we point to related literature in empirical exploration of the neural network energy landscapes and previous applications of the NEB method to neural networks.
%Then, in \cref{sec:neb}, we present the Nudged Elastic Band method in the detail necessary for this application.
%We present our experiments in \cref{sec:experiments} before we sum up our work in \cref{sec:conclusion}.

\section{Related Work}
\label{sec:related}

In discussions about why neural networks generalise despite the extremely large number of parameters, one often finds the argument that wide minima generalise better~\cite{keskar16large}.
This picture is confirmed when visualising the parameter space on a random plane around a minimum~\cite{li17visualizing}.
% However, $l^2$-regularisation breaks this symmetry.
% This contradiction is solved by locally estimating the entropy around a minimum to define the sharpness of the valley~\cite{chaudhari16entropy}.
We draw a completely different image of the loss landscape:
Minima are not located in finite-width valleys, but there are paths through the parameter space along which the loss remains very close to the value at the minima.
A similar view had previously been conjectured by \cite{sagun17empirical}.
They find flat linear paths between minima that are close in parameter space by construction.
We extend their work by constructing flat paths between arbitrary minima.

It has previously been shown that minima of networks with ReLU activations are degenerate~\cite{dinh17sharp}:
One can scale all parameters in one layer by a constant $\alpha$ and in following layer by $\alpha^{-1}$ without changing the output of the network.
Here, we provide evidence for a different kind of degeneracy:
We construct paths between independent minima that are essentially flat.

\cite{freeman16topology} showed that local minima are connected without large barriers for a CNN on MNIST and an RNN on PTB next word prediction.
On CIFAR10, they found a more frustrated landscape.
We extend their work in two ways:
First, we consider ResNets and DenseNets that outperform plain CNNs by a large margin.
Second, we apply a state of the art method for connecting minima from molecular statistical mechanics:
%The characterisation of energy surfaces by connecting minima through low-energy paths originates from molecular statistical mechanics.
%Here, the parameters of the energy function are three-dimensional coordinates of the atoms and molecules in a system.
%A path with low energy can be used to define reaction coordinates for chemical reactions and estimating the rate at which a reaction occurs~\cite{wales98archetypal}.
The Automated Nudged Elastic Band (AutoNEB) algorithm~\cite{kolsbjerg16automated} which in turn is based on the Nudged Elastic Band (NEB) algorithm~\cite{jonsson1998nudged}.
We additionally systematically replace paths that contain relatively high loss barriers.
Combining the above we find paths with essentially no energy barrier.

NEB has so far been applied to a multi-layer perceptron with a single hidden layer~\cite{ballard16energy}.
High energy barriers between the minima of network were found when using three hidden neurons, and disappeared upon adding more neurons to the hidden layer.
In follow-up work,~\cite{ballard17perspective} trained a multi-layer perceptron with a single hidden layer on MNIST.
They found that with $l^2$-regularisation, the landscape had no significant energy barriers.
However, for their network they report an error rate of $14.8\%$ which is higher than the $12\%$ achieved even by a linear classifier~\cite{lecun98gradient} and the 0.35\% achieved with a standard CNN~\cite{ciresan11flexible}.
%They studied the effect of regularisation on this landscape and found that the energy in the training landscape.
%\cite{gamst17energy} produced an artificial dataset same architecture. They don't recover the ground truth.
%\todo{The second one is a really shady reference.}

In this work, we apply AutoNEB to a nontrivial network for the first time, and make the surprising observation that different minima of state of the art networks on CIFAR10 and CIFAR100 are connected through essentially flat paths.

After submission of this work to the International Machine Learning Conference (ICML) 2018, \cite{garipov18loss} independently reported that they also constructed paths between neural network minima.
They study the loss landscape of several architectures on CIFAR10 and CIFAR100 and report the same surprising observation:
minima are connected by paths with constantly low loss.

\section{Method}
\label{sec:neb}

In the following, we use the terms \textit{energy} and \textit{loss} interchangeably.

\subsection{Minimum Energy Path}

A neural network loss function depends on the architecture, the training set and the network parameters $\theta$.
Keeping the former two fixed, we simply write $L(\theta)$ and start with two parameter sets $\theta_1$ and $\theta_2$.
In our case, they are minima of the loss function, i.e.~they result from training the networks to convergence.
The goal is to find the continuous path $p^*$ from $\theta_1$ to $\theta_2$ through parameter space with the lowest maximum loss:
\begin{align*}
	p(\theta_1, \theta_2)^* = \argmin_{p \text{ from } \theta_1 \text{ to } \theta_2} \big\{ \max_{\theta \in p} L(\theta) \big\}.
\end{align*}
%Continuous means that we can sample the path to arbitrary precision and find no higher loss value than $L_t(p)$.
For this optimisation to be tractable, the loss function must be sufficiently smooth, i.e.~contain no jumps along the path.
The output and loss of neural networks are continuous functions of the parameters~\cite{montufar14number}; only the derivative is discontinuous for the case of ReLU activations.
However, we cannot give any bounds on how steep the loss function may be.
%While advances are made in what distribution of slopes can be expected, this does not give us a tight upper bound \todo{Cite someone}.
We address this problem by sampling all paths very densely.

Such a lowest path $p^*$ is called the \textit{minimum energy path} (MEP)~\cite{jonsson1998nudged}.
We refer to the parameter set with the maximum loss on a path as the ``saddle point'' of the path because it is a true saddle point of the loss function.

%One special path among all possible MEPs is one where the gradient of the loss perpendicular to the path is zero.
%\cref{fig:min_ene_path} shows such a MEP.

%Minimum energy paths between several minima can be summarised in a disconnectivity graph~\cite{wales98archetypal}.
%This is a form of dendrogram where two clusters are merged when the .

%When, for a given system, many such paths are found between a variety of minima, a network.
%This can the be used to analyse the 

%In this work, we compute approximations to minimum energy paths for recent deep neural networks on image classification \todo{and segmentation} datasets.

In low-dimensional spaces, it is easy to construct the exact minimum energy path between two minima, for example by using dynamic programming on a densely sampled grid.
%(Kruskal-type region growing until regions collide).

This is not possible for present day's neural networks with parameter spaces that have millions of dimensions.
We thus must resort to methods that construct an approximation of the MEP between two points using some local heuristics.
In particular, we resort to the Automated Nudged Elastic Band (AutoNEB) algorithm~\cite{kolsbjerg16automated}.
This method is based on the Nudged Elastic Band (NEB) algorithm~\cite{jonsson1998nudged}.

NEB bends a straight line segment by applying gradient forces until there are no more gradients perpendicular to the path.
Then, as for the MEP, the highest point of the resulting path is a critical point.
While this critical point is not necessarily the saddle point we were looking for, it gives an upper bound for the energy at the saddle point.

In the following, we present the mechanical model behind and the details of NEB.
We then proceed to AutoNEB.
%In the classical NEB literature, the parameter sets that sample the path are called \textit{images}.
%This is misleading in the context of computer vision and we refer to them as \textit{pivots} instead.

%A continuous version of the path found using NEB is given by linearly interpolating between subsequent states.

%The start point of the iteration is some initial guess of the path. When no further information is available, the linear interpolation is used

%For a molecular system, a linear initialisation can lead to large forces since particles can come too close to each.
%Such restrictions are not known for neural networks.
%\todo{We \textit{do} notice that the linear interpolation between minima leads to worse-than-random losses.}

%In the following, we recapture the version of NED

\paragraph{Mechanical Model}
A chain of $N+2$ pivots (parameter sets) $p_i $ for $i = 0, \dots, N+1$ is connected via springs of stiffness $k$.
The initial and the final pivots are fixed to the minima to connect, i.e. $p_0 = \theta_1$ and $p_{N+1} = \theta_2$.
Using gradient descent, the path that minimises the following energy function is found:
\begin{align}
	\label{eq:PEB}
	E(p) = \sum_{i=1}^{N} L(p_i) + \sum_{i=0}^{N} \frac12 k \norm{p_{i+1} - p_i}^2
\end{align}
The problem with this energy formulation lies in the choice of the spring constant:
If, on the one hand, $k$ is too small, the distances between the pivots become larger in areas with high energy.
However, identifying the highest point on the path and its energy is the very goal of the algorithm, so the sampling rate should be high in the high-energy regions.
If, on the other hand, $k$ is chosen too large, it becomes energetically advantageous to shorten and hence straighten the path as the spring energy grows quadratically with the total length of the path.
This cuts into corners of the loss surface and the resulting path can miss the saddle point.
%\todo{This means that we have to use many, many pivots and small $k$.}

\begin{figure}[th]
	\centering
	\includegraphics[width=\linewidth]{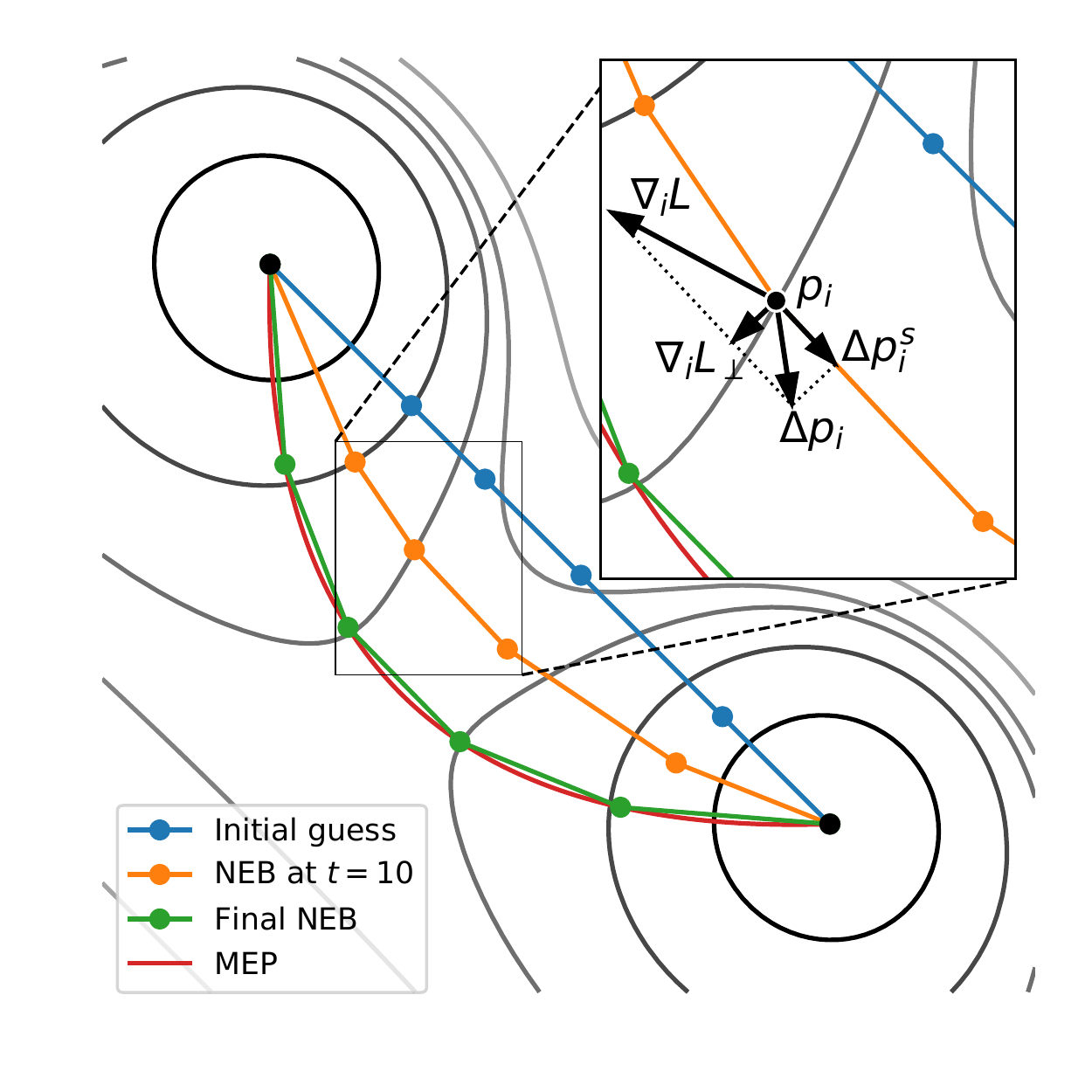}
	\vspace{-1cm}
	\caption{
		Two dimensional loss surface, with two minima connected by a minimum energy path (MEP) and a nudged elastic band (NEB) at iteration 0, 10 and converged.
		Construction of NEB update $\Delta p_i$ for one pivot.
		The tangent points to the neighbouring pivot with higher energy.
		Re-distribution $\Delta p_i^s$ acts parallelly and the loss force $\nabla_i L$ perpendicularly to the tangent.
	}
	\label{fig:all_forces}
\end{figure}

\paragraph{Nudged Elastic Band}
Inspired by the above model,~\cite{jonsson1998nudged} presented the \textit{Nudged Elastic Band} (NEB).
For brevity, we directly present the improved version by~\cite{henkelman00improved}.
The force resulting from \cref{eq:PEB} consists of a force derived from the loss and a force originating from the springs:
\begin{align*}
	F_i = - \nabla_{p_i} E(p) = F_i^L + F_i^S
\end{align*}
For NEB, the physical forces are modified, or \textit{nudged}, so that the loss force only acts perpendicularly to the path and the spring force only parallelly to the path (see also \cref{fig:all_forces}):
\begin{align*}
	F_i^{\text{NEB}} &= F_i^L \big|_\perp + F_i^S \big|_\parallel.
\end{align*}
The direction of the path is defined by the local tangent $\hat\tau_i$ to the path.
The two forces now read:
\begin{align}
	\label{eq:nudged_forces_details}
	\begin{split}
		F_i^L \big|_\perp &= -(\nabla L(p_i) - (\nabla L(p_i) \cdot \hat\tau_i) \hat\tau_{i}) \\
		F_i^S \big|_\parallel &= (F_i^S \cdot \hat\tau_i) \hat\tau_i
	\end{split}
\end{align}
where the spring force opposes unequal distances along the path:
\begin{align}
	\label{eq:spring}
	F_i^S = - k (\norm{p_i - p_{i - 1}} - \norm{p_{i + 1} - p_i})
\end{align}

In this formulation, high energy pivots no longer ``slide down'' from the saddle point. 
The spring force only re-distributes pivots on the path, but does not straighten it.
Pivots can be spaced unequally by introducing target distances or unequal spring constants into \cref{eq:spring}.

The local tangent is chosen to point in the direction of one of the adjacent pivots ($\mathcal{N}$ normalises to length one):
\begin{align*}
	\hat\tau_i = \mathcal{N} \begin{cases}
		p_{i+1} - p_i & \text{if } L(p_{i+1}) > L(p_{i-1}) \\
		p_i - p_{i-1} & \text{else.}
	\end{cases}
\end{align*}
This particular choice of $\hat{\tau}$ prevents kinks in the path and ensures a good approximation near the saddle point~\cite{henkelman00improved}.

The above procedure requires the following hyperparameters:
The spring stiffness $k$ and number of pivots $N$.% need to be chosen.

\cite{sheppard08optimization} claim that a wide range of $k$ leads to the same result on a given loss surface.
However, if chosen too large, the optimisation can become unstable.
If it is too small, an excessive number of iterations are needed before the pivots become equally distributed.
We did not find a value for $k$ that worked well across different loss surfaces and number of pivots $N$.
Instead, we re-distribute the pivots in each iteration $t$ and set the actual spring force to zero.
The loss force is still restricted to act parallelly to the path.
In the literature, this is sometimes referred to as the \textit{string method}~\cite{sheppard08optimization}.

\cref{alg:neb} shows how the initial path is iteratively updated using the above forces.
As a companion, \cref{fig:all_forces} visualises the forces in one update step for a two dimensional example.
In this formulation, we use gradient descent to update the path.
Any other gradient based optimiser can be used.
It typically introduces additional hyperparameters, for example a learning rate $\gamma$.
The number of iterations~$T$ should be chosen large enough for the optimisation to converge.

\begin{algorithm}[htb]
	\caption{NEB}
	\label{alg:neb}
	\begin{algorithmic}
		\STATE {\bfseries Input:} initial path $p^{(0)}$ with $N+2$ pivots, \\~~~~$p_0^{(0)} = \theta_1$ and $p_{N+1}^{(0)} = \theta_2$.
		\FOR{t = $1, \dots, T$}
			\STATE Redistribute pivots on path $p^{(t - 1)}$ and store as $p$.
			\FOR{i = $1, \dots, N$}
				\STATE Compute projected loss force $F_i = F^L_i\big|_\perp$.
				\STATE Store pivot $p_i^{(t)} = p_i + \gamma F_i$.
			\ENDFOR
		\ENDFOR
		\STATE {\bfseries return} final path $p^{(T)}$
	\end{algorithmic}
\end{algorithm}

The evaluation time of \cref{alg:neb} rises linearly with the number of iterations and the number of pivots on the path.
Computing the NEB forces can trivially be parallelised over the pivots.

The number of pivots $N$ trades off between computational effort on the one hand and subsampling artefacts on the other hand.
%For molecular systems, the distance between two pivots is intrinsically given by the system (e.g. by the distances between particles).
In neural networks, it is not known what sampling density is needed for traversing the parameter space.
We use an adaptive procedure that inserts more pivots where needed:

\paragraph{AutoNEB}
The Automated Nudged Elastic Band (AutoNEB, \cref{alg:min_connect}) wraps the above NEB algorithm~\cite{kolsbjerg16automated}.
It runs NEB only for a small number of iterations $T$ at a time, initially with a small number of pivots $N$.
It is then checked if the current pivots accurately sample the path.
If sampling is not dense enough, new pivots are added at locations where it is estimated that the path requires more accuracy, see \cref{app:insertion}.
This procedure is repeated several times.
%New pivots are inserted where the true loss values deviate from the linear interpolation between each neighbouring pivot pair larger than a certain threshold, as explained in \cref{app:insertion}.
%and requires the redistribution of pivots in \cref{alg:neb} to handle unequal spaces between pivots.

\begin{algorithm}[htb]
	\caption{AutoNEB}
	\label{alg:min_connect}
	\begin{algorithmic}
		\STATE {\bfseries Input:} Minima to connect $\theta_1, \theta_2$.
		\STATE Initialise $N$ pivots equally spaced on line segment $(\theta_1, \theta_2)$.
		\FOR{$t' = 1, \dots, T'$}
			\STATE Optimise path using NEB (see \cref{alg:neb}).
			\STATE Evaluate loss along NEB.
			\STATE Insert pivots where residuum is large.
		\ENDFOR
		\STATE {\bfseries return} path after final iteration.
	\end{algorithmic}
\end{algorithm}

\subsection{Local minimum energy paths}

AutoNEB is not guaranteed to find the true MEP.
Instead, it can get stuck in local minimum energy paths (local MEPs) with spuriously high saddle point losses.
The good news is that the graph of minima and local MEPs has an ultrametric property:
Suppose some local MEPs from a minimum $A$ to $B$ and from $B$ to $C$ are known.
We call them $p_{AB}$ and $p_{BC}$.
The respective saddle point energies give an upper bound for the true saddle point energies (marked with an asterisk):
\begin{align*}
	L_{AB}^* &\le L_{AB} = \max_{\theta \in p_{AB}} L(\theta) \\
	L_{BC}^* &\le L_{BC} = \max_{\theta \in p_{BC}} L(\theta)
\end{align*}
The concatenation of the two paths yields an upper bound for the true saddle point energy between $A$ and~$C$ (ultrametric triangle inequality):
\begin{align*}
	L_{AC}^* \le \max\{ L_{AB}, L_{BC} \}
\end{align*}

\begin{figure}[t]
	\centering
	\includegraphics[width=\linewidth]{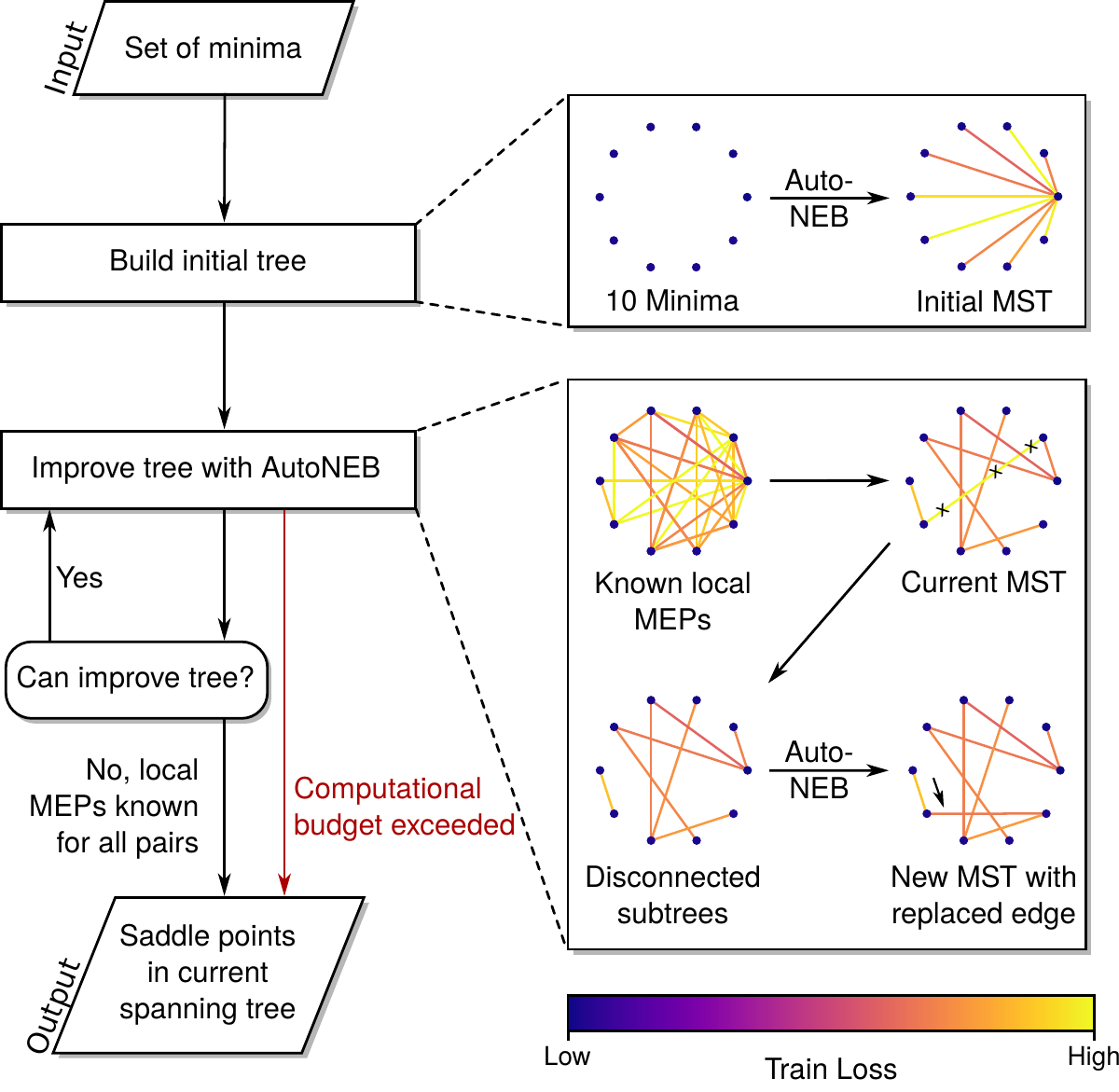}
	\vspace{-.5cm}
	\caption{
		Overview over \cref{alg:choose_pairs} and examples for the first nine iterations and some later iteration.
		First, all minima are connected to one particular minimum.
		Then, AutoNEB computes new local MEPs to circumvent the worst local MEPs in the minimum spanning tree.
		This is repeated until local MEPs are known between all pairs of minima or the procedure is stopped early.
		Whenever the algorithm stops, an upper bound for each pair of minima is available via the minimum spanning tree.
	}
	\label{fig:transition_network}
\end{figure}

\begin{proof}
	Concatenating the paths $p_{AB}$ and $p_{BC}$ gives a new path $p_{AC}$ connecting $A$ to~$C$.
	The saddle point is located at the maximum loss along a path and hence the saddle point energy of $p_{AC}$ is $L_{AC} = \max\{ L_{AB}, L_{BC} \}$.
\end{proof}

This has three consequences:
\begin{enumerate}
	\item As soon as the minima and computed local MEPs form one connected graph, upper bounds for all saddle energies are available.
	We can hence very quickly get upper bounds for all pairs of minima by connecting one minimum to all others.
	\item When AutoNEB finds a bad local MEP, this can be addressed by computing paths between other pairs of minima.
	As soon as a lower path is found by concatenating other paths, the bad local MEP can be removed.
	This means that the bad local paths can easily be corrected for.
	\item When we evaluate the saddle point energies of a set of computed local MEPs, we can ignore paths with higher energy than the concatenation of paths with a lower maximal energy. \\
	These lowest local MEPs form a minimum spanning tree in the available graph~\cite{gower69minimum}.
	A Minimum Spanning Tree (MST) can be found efficiently, e.g. using Kruskal's algorithm.
\end{enumerate}

We resort to a heuristic (\cref{fig:transition_network}, \cref{alg:choose_pairs}) to systematically sample edge costs from a latent graph to find or approximate its MST.
Since running AutoNEB is computationally expensive (comparable to training the corresponding network once), we stop the iteration when the lightened spanning tree found so far contains only similar saddle point energies.

\begin{algorithm}[th]
	\caption{Energy Landscape Exploration}
	\label{alg:choose_pairs}
	\begin{algorithmic}
		\STATE {\bfseries Input:} set of minima $\theta_i$.
		\STATE Connect $\theta_1$ to all $\theta_i, i \neq 1$, yielding a spanning tree.
		\REPEAT
		\STATE Remove edge $p_o$ with highest loss from spanning tree.
		%\STATE The tree splits into two disconnected subtrees.
		\STATE From each resulting tree, try to select one minimum, \\~~~~so that no local MEP is known for the pair.
		\IF{search failed}
		\STATE Re-insert $p_o$ and ignore it when searching for the\\~~~~highest edge in the future.
		\ELSE
		\STATE Compute new path $p_n$ using AutoNEB.
		\IF{$L_{p_n} < L_{p_o}$}
		\STATE Add $p_n$ to the tree, making tree ``lighter''.
		\ELSE
		\STATE Re-insert $p_o$ to the tree (no better path was found).
		\ENDIF
		\ENDIF
		\UNTIL{one local MEP is known for each pair of minima\\~~~~or computational budget is exceeded.}
		\STATE {\bfseries return} saddle points in minimum spanning tree.
	\end{algorithmic}
\end{algorithm}

\section{Experiments}
\label{sec:experiments}

We connect minima of different CNNs, ResNets~\cite{he16deep} and DenseNets~\cite{huang17densely} on the image classification tasks CIFAR10 and CIFAR100~\cite{krizhevsky09learning} using AutoNEB.
Per architecture, we consider ten minima.\footnote{Source code is available at \url{https://github.com/fdraxler/PyTorch-AutoNEB}.}

The minima are constructed from multiple random initialisations and are truly distinct:
On the test data, the set of misclassified images differs between the minima.
More precisely, on the ResNet and DenseNet architectures, we observe a maximum 70\% overlap of the samples that are misclassified at two minima, proving their distinctiveness.

We report the average cross-entropy loss and misclassification rates over the full training and test data for the minima found.
For the final evaluation, we reduce the saddle points to the minimum spanning tree with the saddle training loss as weight.

\begin{figure*}[ht]
	\centering
	\includegraphics[width=\linewidth]{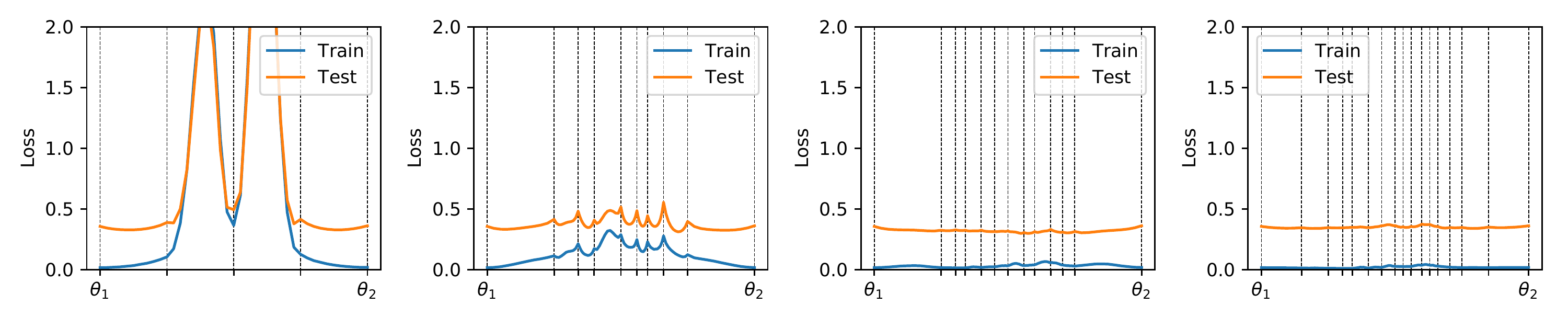}
	\vspace{-1cm}
	\caption{
		Typical snapshots of the loss along a path connecting two minima, as pivots are inserted into a nudged elastic band:
		(1) After the first cycle, typically one or two corners are cut.
		New pivots are inserted at high loss values, here between the second and the third pivot.
		(2) After four cycles of high learning rate, the highest loss on the path is reduced by a factor of five.
		Between pivots we find low energy regions that we attribute to the high learning rate of $0.1$.
		(3) The first round with low learning rate of $0.01$ reduces the energy by another factor of two.
		(4) After 14 cycles, no major energy bumps exist between the pivots, the procedure is converged.
		Results shown are for a ResNet-20 on CIFAR10.
	}
	\label{fig:neb_cycles}
\end{figure*}

\subsection{AutoNEB schedule}

The set-up is identical for all network architectures, except for the batch sizes which we note in each case.

The minimum pairs to connect are ordered by \cref{alg:choose_pairs}.
For each minimum pair, AutoNEB (see \cref{alg:min_connect}) is run for a total of 14 cycles of NEB.
The loss is evaluated for each pivot on a random batch.

After each cycle, new pivots are inserted at positions where the loss exceeds the energy estimated by linear interpolation between pivots by at least 20\% compared to the total energy difference along the path.
Comparing to the total loss difference prioritises big errors which is beneficial as each additional pivot implies one more loss evaluations per iteration.
The energy is evaluated on nine points between each pair of neighbouring pivots.
%Evaluating the energies after each cycle takes approximately half the time compared to running a corresponding cycle of 1000 steps.

As optimiser, we use SGD with momentum $0.9$ and $l^2$-regularisation with $\lambda = 0.0001$.

The NEB cycles are configured with a learning rate decay:
\begin{enumerate}
	\item Four cycles of 1000 steps each with learning rate 0.1.
	\item Two cycles with 2000 steps and learning rate 0.1.
	The number of steps was increased as it did not prove necessary inserting new pivots after 1000 steps.
	\item Four cycles of 1000 steps with learning rate 0.01.
	The loss drops significantly in this phase.
	\item No big improvement was seen in the last four cycles of 1000 steps each with a learning rate of 0.001.
\end{enumerate}
\cref{fig:neb_cycles} shows typical snapshots of the loss-along-path between the above cycles.

\subsection{Architectures}

We consider a wide range of architectures, from shallow CNNs to recent deep networks with skip connections.

\paragraph{Basic CNN}
We analyse CNNs without skip connections with a variety of depths and widths on both CIFAR10 and CIFAR100.
We name them ``CNN-$W \PLH D$'' where $W$ corresponds to the width of each layer (number of channels) and $D$ to the number of convolutional layers.
Each convolution is $5\times5$, a max pooling layer of 2 is attached to each convolution, and a single hidden fully connected layer of width 256 and batch normalisation \cite{ioffe15batch} are used.
We consider the one-layer CNN-$12 \PLH 1$, CNN-$24 \PLH 1$, CNN-$36 \PLH 1$, CNN-$48 \PLH 1$ and CNN-$96 \PLH 1$, and the multi-layer CNN-$48 \PLH 2$ and CNN-$48 \PLH 3$.

\paragraph{ResNet}
We train ResNets on both CIFAR10 and CIFAR100 (ResNet-20, -32, -44 and -56) following the training procedure in~\cite{he16deep}.
For ResNet-20 and ResNet-32, the best local MEPs were found using a batch size of 512 training samples.
For ResNet-44 and ResNet-56, this number was decreased to 256.

\paragraph{DenseNet}
We train a DenseNet-40-12 and a DenseNet-100-12-BC on both CIFAR10 and CIFAR100 following the training procedure in~\cite{huang17densely}.
The AutoNEB batch size was set to 256.

\subsection{Saddle point losses}

\begin{figure}[ht]
	\centering
	\includegraphics[width=\linewidth]{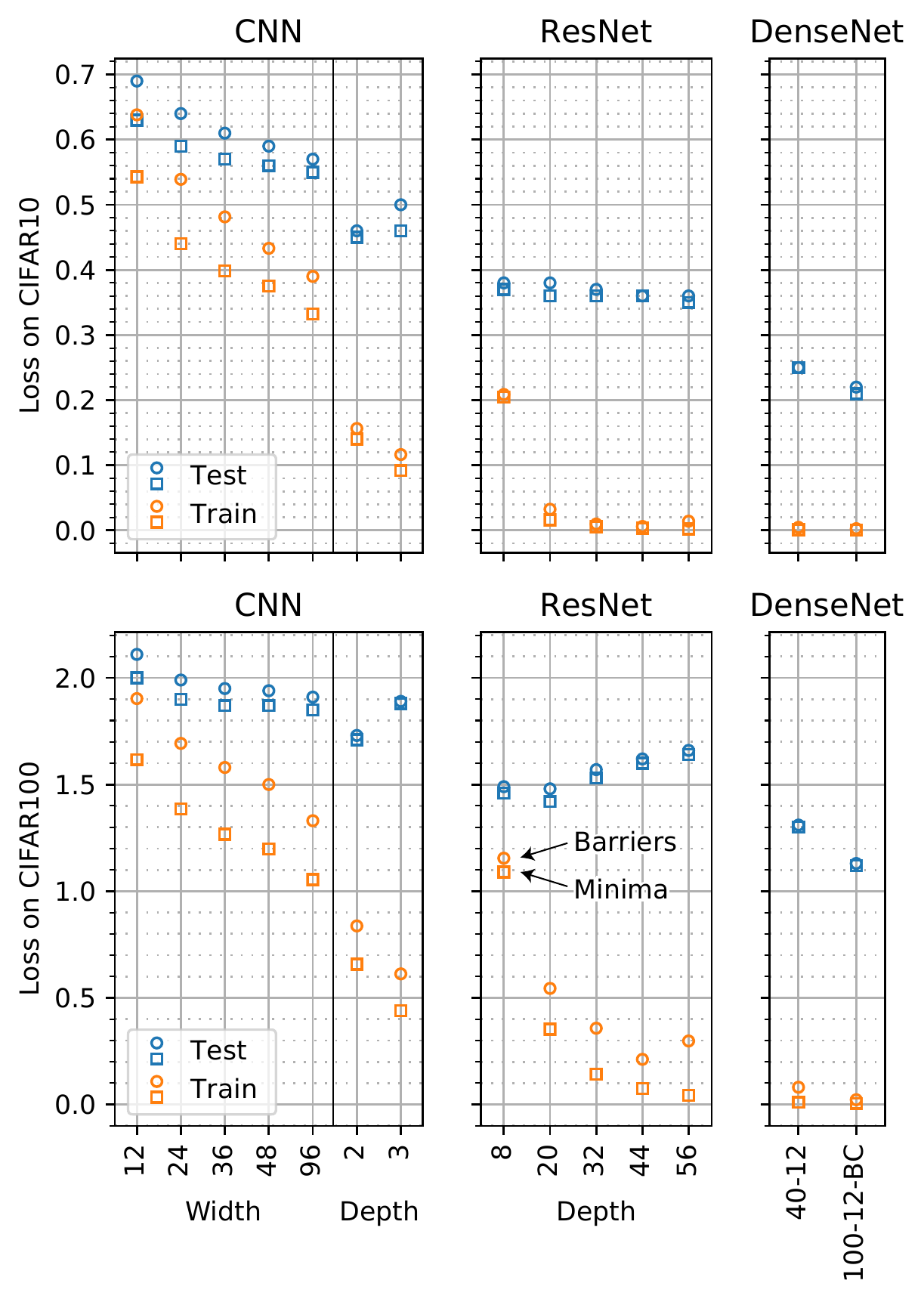}
	\vspace{-.7cm}
	\caption{
		Core result:
		Minimum ($ \blacksquare $) and saddle point ($ \bullet $) losses on training and test set.
		The training (orange) saddle losses are almost identical to the minima on CIFAR10 on the deep architectures (middle and right).
		On CIFAR100, a small gap remains that vanishes in comparison to the loss of a randomly initialised network.
		On a simple architecture (left), increasing width and especially depth closes the gap between the barriers and the minima.
		On the test set (blue), the points of the minima and the saddle points are very close.
		Error bars are omitted since they would be smaller than the markers.
	}
	\label{fig:summary}
\end{figure}

The saddle point losses for both training and test sets found by AutoNEB are shown in \cref{fig:summary}, and listed in detail in \cref{tab:summary} in \cref{app:values}.
They are small for the shallow networks and almost negligible for the deep residual networks.

Compare the saddle point loss to the loss at the minima on the training and on the test set.
For the shallow CNNs on the one hand, the saddle loss is found quite close to the test loss.
On the other hand, the saddle loss of the ResNets and DenseNets lies very close to the training loss.

\begin{figure}[th]
	\centering
	\includegraphics[width=\linewidth]{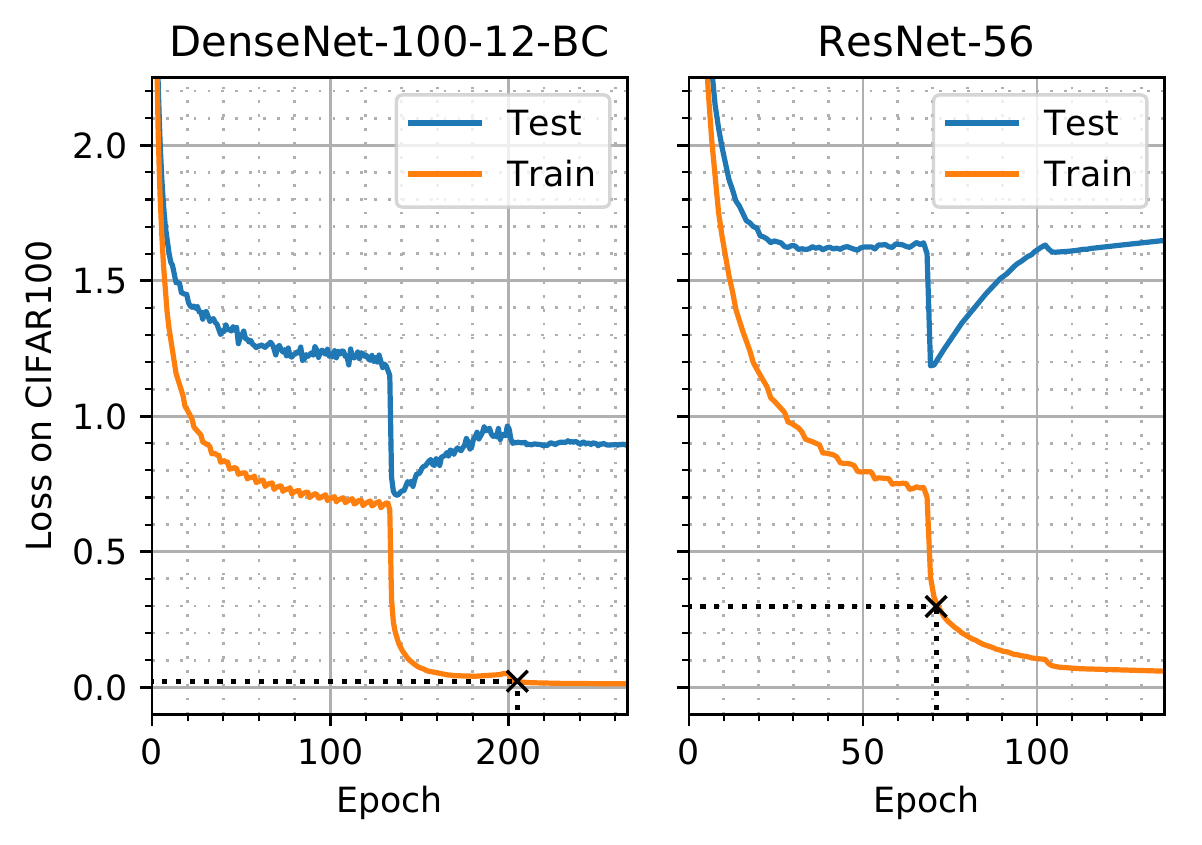}
	\vspace{-1cm}
	\caption{
		Learning curves illustrating how late in training the loss drops below the saddles.
		The training loss passes the mean saddle point energy at about 205 (72) epochs or 78\% (52\%) of training.
		Among all deep architectures considered, the average saddle point crosses the training loss last for DenseNet-100-12-BC and earliest for ResNet-56, both on CIFAR100.
		The crossing for the DenseNet had to be identified in a log plot.
	}
	\label{fig:first_passage}
\end{figure}

Further, we measure how late during training the learning curve crosses the saddle loss, as visualised in \cref{fig:first_passage}.
The learning curve falls below the saddle point energy only after the first learning rate decay and an additional significant drop of the loss for all architectures.
For the wider CNNs on CIFAR10 and the majority of ResNets and DenseNets, the losses meet even after the second decay, i.e.~in the final phase of learning.

We observe the following trend:
The \textit{deeper} and \textit{wider} an architecture, the \textit{lower} are the saddles between minima until they essentially \textit{vanish for current-day deep architectures}.
The more complex dataset CIFAR100 raises the barriers.

At the same time, the test accuracy is preserved:
The classification error only increases slightly by maximally $0.5\%$ ($2.2\%$) for all deep architectures on CIFAR10 (CIFAR100) compared to the minima.

We conclude that the saddle points have surprisingly low loss with respect to the metrics above.
In other words, there are essentially no loss barriers in current-day deep architectures.
%The deeper and wider an architecture is, the lower are the barriers in its loss surface.

\subsection{Properties of obtained local MEPs}

\begin{figure}[hbt]
	\centering
	\includegraphics[width=\linewidth]{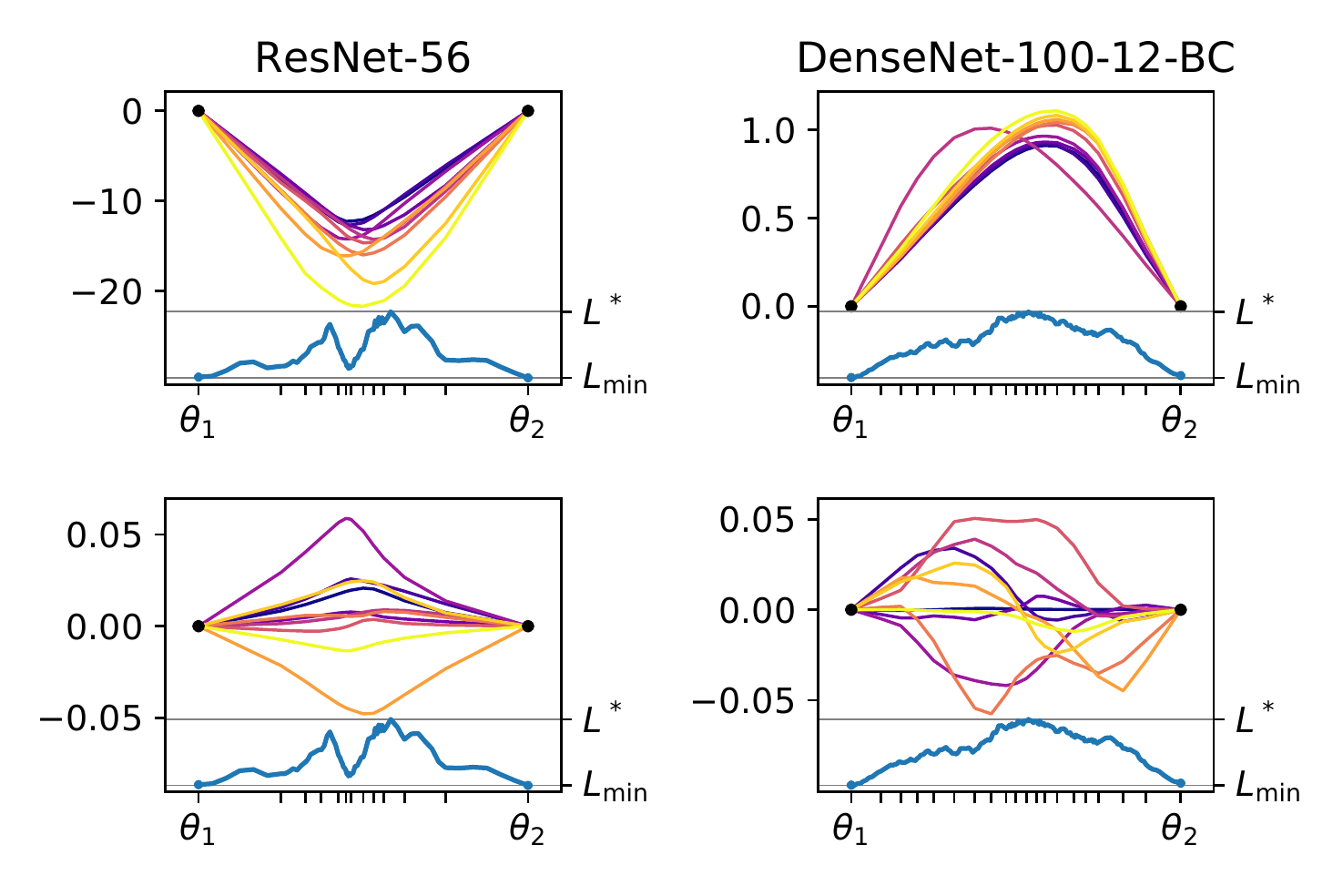}
	\vspace{-1cm}
	\caption{
		Network parameters follow a smooth trajectory along the MEPs.
		For each coordinate, the distance to the linear interpolation is shown (negative means closer to origin).
		The largest deviations from the linear path occur near the saddle point.
		The upper plots show those ten coordinates that deviate most from the linear interpolation.
		The ten coordinates in the lower row are chosen randomly.
		Also shown is the training loss along the two paths.
		The dataset is CIFAR100.
	}
	\label{fig:parallel_corods}
\end{figure}

The local MEPs between the minima not only have very low loss, they also follow simple trajectories.
\cref{fig:parallel_corods} shows some coordinates of two local MEPs in a parallel coordinate plot.
We find that each coordinate has a smooth path.
The largest deviations occur near the saddle point of the path.
The paths are between $ 50\% $ to $ 2.5 $ times longer than the direct connection between the minima.

\section{Discussion}

We have pointed out an intriguing property of the loss surface of current-day deep networks, by upper-bounding the saddle points between the parameter sets that result from stochastic gradient descent, a.k.a. ``minima''.
These empirical upper bounds are astonishingly close to the loss at the minima themselves.
The experiments on the CNNs suggest that the disappearance of barriers emerges as the networks get wider and especially deeper.
At this point, we cannot give a formal characterization of the regime in which this finding holds.
A formal proof is also complicated by the fact that the loss surface is a function not only of the parameters and the architecture, but also of the training set; and the distribution of real-world structured data such as images or sentences does not lend itself to a compact mathematical representation.
That said, we want to make two related arguments that may help explain why we observe no substantial barrier between minima.

\subsection{Resilience}

State of the art neural networks have dozens or hundreds of neurons / channels per layer, and skip connections between non-adjacent layers.
Assume that by training, a parameter set with low loss has been identified.
Now if we perturb a single parameter, say by adding a small constant, but leave the others free to adapt to this change to still minimise the loss, it may be argued that by adjusting somewhat, the myriad other parameters can ``make up'' for the change imposed on only one of them.
After this relaxation, the procedure and argument can be repeated, though possibly with the perturbation of a different parameter. % than in the previous rounds.

This type of resilience is exploited and encouraged by procedures such as Dropout~\cite{srivastava14dropout} or ensembling~\cite{hansen90neural}.
It is also the reason why neural networks can be greatly condensed before a substantial increase in loss occurs~\cite{liu17learning}. 

\subsection{Redundancy}

\begin{figure}[hbt]
	\centering
	\vspace{0.175cm}
	\includegraphics[width=\linewidth]{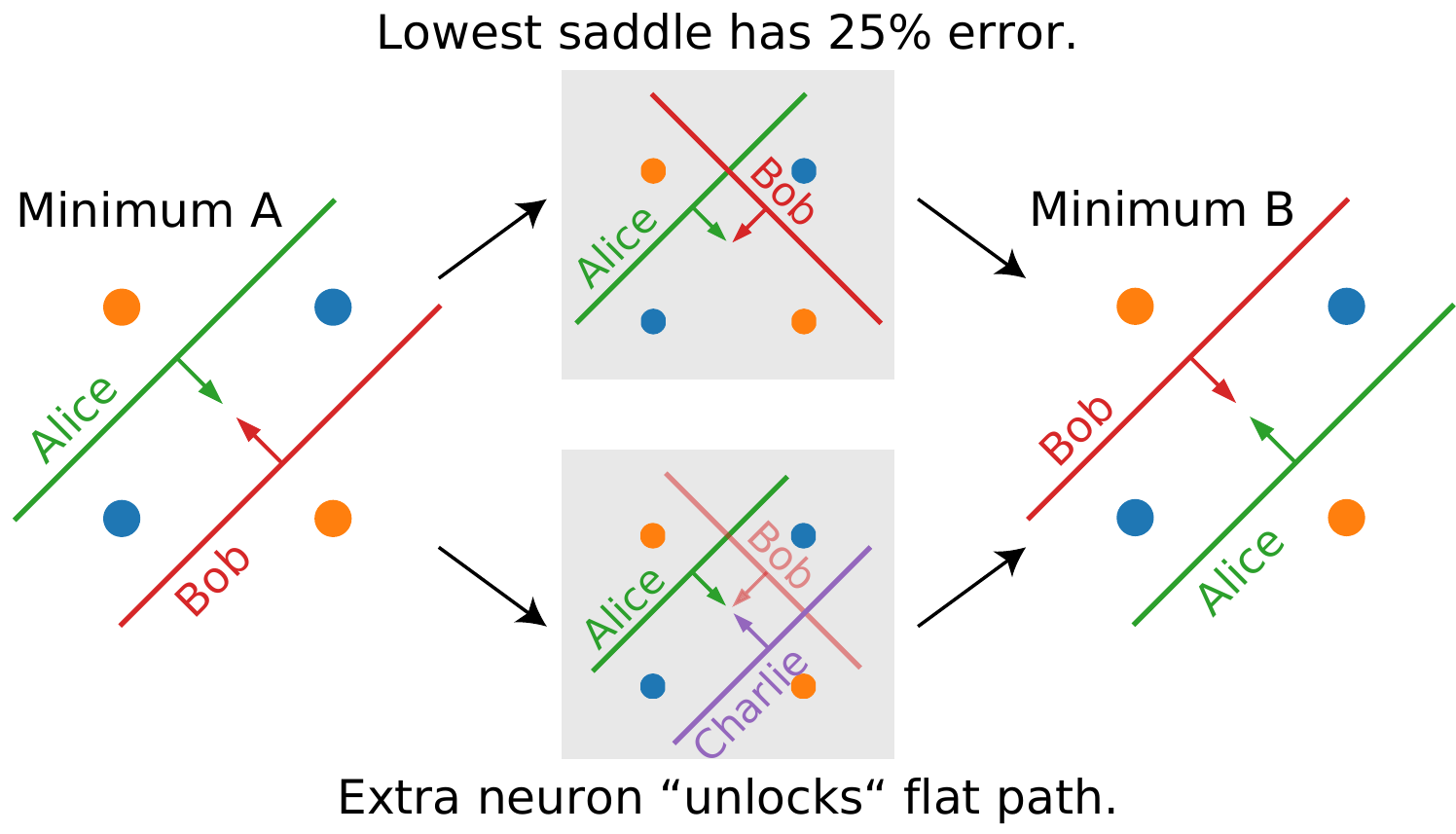}
	\vspace{-0.4cm}
	\caption{Network capacity for XOR dataset: The continuous transition from one minimum \textit{(left)} to another minimum \textit{(right)} is not possible without misclassifying at least one instance \textit{(upper middle)}. \textit{(Lower middle)} Adding one helper neuron makes the transition possible while always predicting the right class for all data points, i.e.~by turning off the outgoing weight of Bob.}
	\label{fig:xor}
\end{figure}

Consider the textbook example of a two-layer perceptron that can fit the XOR problem.
The two neurons traditionally used in the first hidden layer -- let's call them Alice and Bob -- are shown in \cref{fig:xor} on the left.
We can obtain an equivalent network by exchanging Alice and Bob (and permuting the weights of the neuron in the second hidden layer, not shown).
This network, also corresponding to a minimum of the loss surface, is shown in \cref{fig:xor} on the right.
Now, any path between these two minima will entail parameter sets such as the one in the upper centre of \cref{fig:xor} that incur high loss.

If, on the other hand, we introduce an auxiliary neuron, Charlie, we can play a small choreography:
Enter Charlie.
Charlie stands in for Bob.
Bob transitions to Alice's role via \cref{fig:xor}, lower centre.
Alice takes over from Charlie.
Exit Charlie.
If the neuron in the second hidden layer adjusts its weights so as to disregard the output from the neuron-in-transition, the entire network incurs no higher loss than at the two original minima.

We have constructed a perfect minimum energy path through increasing the \textit{width}.
Similarly, it is possible to construct a zero-loss path by adding a second two-neuron layer to the network, that is by increasing the \textit{depth} of the network.

\section{Conclusion}
\label{sec:conclusion}

We find that the loss surface of deep neural networks contains paths with constantly low loss.
The paths connect the minima so that they form one single connected component in the loss landscape.
The barriers are especially low with increasing depth and width.

We put forth two closely related explanations in the above.
Both hold only if the network has some extra capacity, or degrees of freedom, to spare.
Empirically, this seems to be the case for modern-day architectures applied to standard problems.

This has the profound implication that low Hessian eigenvalues exist apart from the eigenvectors with analytically zero eigenvalues due to scaling.

We introduce AutoNEB for the characterisation of current-day architectures for the first time.
The method opens the door to further empirical research on the energy landscape of neural networks.
When the hyperparameters of AutoNEB are further refined, we expect to find even lower paths up to the level where the true saddle points are recovered.
It is then interesting to see if certain minima have a higher barrier between them than others.
This makes it possible to recursively form clusters of minima, i.e.~using single-linkage clustering.
In the traditional energy landscape literature, this kind of clustering is summarised in disconnectivity graphs~\cite{wales98archetypal} which can help visualise very high-dimensional surfaces.

On the practical side, we envisage using the resulting paths as a large ensemble of neural networks \cite{garipov18loss}, especially given that we observe marginally lower test loss along the path.

More importantly, we hope these observations will stimulate new theoretical work to better understand the nature of the loss surface, and why local optimisation on such surfaces results in networks that generalize so well.

\section*{Acknowledgements}

FAH gratefully acknowledges support by DFG under grant no. HA 4364/9-1. 

\bibliography{../Paper/Database}
\bibliographystyle{icml2018}

\clearpage
\newpage
% !TeX root = paper.tex

\appendix    
\appendixpage

\counterwithin{figure}{section}
\counterwithin{table}{section}

\section{AutoNEB insertion}
\label{app:insertion}

\begin{figure}[ht]
	\centering
	\includegraphics[width=\linewidth]{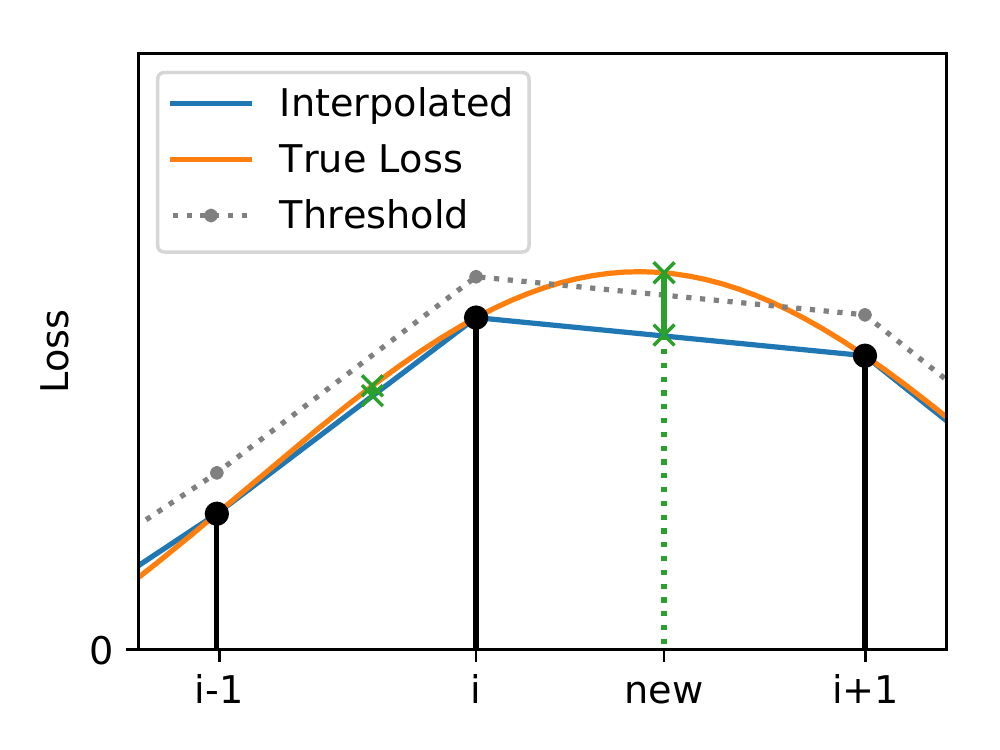}
	\caption{New items are inserted in each cycle of AutoNEB when the true energy at an interpolated position between two points rises too high compared to the interpolated energy. Between $i$ and $i+1$, a new pivot is inserted. Between $i-1$ and $i$, the difference is small enough that no additional pivot is needed.}
	\label{fig:auto_neb_insert}
\end{figure}

New pivots are inserted at locations where the loss values resulting from evaluations rise higher than a certain threshold above the the estimate given the adjacent pivots, see \cref{fig:auto_neb_insert}.
More formally, the loss curve between pivots $i$ and $i+1$ can be approximated by interpolating the pivot values:
\begin{align*}
	L_\text{guess}^{i,i+1}(\alpha) &= L(p_i) (1 - \alpha) + L(p_{i+1}) \alpha
\intertext{where $\alpha \in [0,1]$ interpolates between the pivots. The true loss value a the same position is:}
	L^{i,i+1}(\alpha) &= L\big(p_i (1 - \alpha) + p_{i+1} \alpha\big).
\intertext{Denote the difference as:}
	\Delta L^{i,i+1}(\alpha) &= L^{i,i+1}(\alpha) - L_\text{guess}^{i,i+1}(\alpha)
\end{align*}

If the true loss rises too high above the estimated loss, a new pivot should be inserted.
It is beneficial to first insert pivots at the highest residuum:
Each new pivot requires more expensive gradient evaluations.
The deviation between true loss and estimate is evaluated at discrete positions $\alpha \in \mathcal A \subset (0,1), ~ |\mathcal A| < \infty$.
The differences $\Delta L^{i,i+1}(\alpha)$ are normalised to the range of values of $L(p_i)$.
When this normalised deviation rises above a certain threshold $\vartheta$, a new pivot is inserted, i.e. insert a pivot between $i$ and $i+1$ when:
\begin{align}
\label{eq:residual}
\vartheta > \frac{\Delta L^{i,i+1}(\alpha)}{\max_i L(p_i) - \min_i L(p_i)} =: {\Delta l}^{i,i+1}(\alpha)
\end{align}

Only one pivot is inserted per line segment per AutoNEB iteration.
If several $\alpha$ for one line segment fulfil the above condition, only the position with the highest residuum is chosen.
Additionally, the total number of pivots to insert per iteration is limited so that the highest deviations are prioritised.

\section{Quantitative minimum and saddle point losses}
\label{app:values}

\cref{tab:summary} shows the full list of results.
Here, we compare the numbers in detail to characteristic metrics of a neural network, with error margins below $10\%$ for all energy and error rate values:

The \textit{loss of an untrained network} amounts to $-\log(0.1) = 2.3$ on CIFAR10 and $-\log(0.01) = 4.6$ on CIFAR100.\\
The saddle point energies are between $1/3$ to $1/20$ of the initial loss for the shallow networks (all CNNs and ResNet-8) and \emph{about two orders of magnitude smaller} for the deep residual networks.

The \textit{test loss} at the saddle points of the smallest one-layer CNNs comes close to the test loss of the minima.
The wider and especially deeper the CNN, the closer the saddle loss approaches the minima.
The saddle point energies of the deep residual networks (not ResNet-8) on CIFAR10 are about one order of magnitude smaller than the average minimum loss on the test set.
On CIFAR100, the saddle point energies of the ResNets are smaller than a third of the value on the test set.
For the DenseNets, they are \textit{at least one order of magnitude smaller}.

The \textit{training loss} at the saddle points is at most eight times as large as the training loss of the minima.
This ratio, reported as ``factor'' in \cref{tab:summary}, is hard to interpret directly as it can approach zero when the network fits the data perfectly.
Instead, we report that all saddle losses are closer to the training than the test loss for all but the smallest three basic CNNs.
For all deeper architectures, the saddle loss is \textit{much closer to the training than to the test loss}, see \cref{fig:summary}.

The \textit{classification performance} does not decrease significantly along the paths.
On the ResNets, the error rises by maximally 0.7\% on CIFAR10 and 2.9\% on CIFAR100.
For the DenseNets, the error rises by up to 0.4\% on CIFAR10 and 1.5\% on CIFAR100.
These \textit{differences are small} compared to the error rate at the minima.

\begin{table*}[!b]
	\captionof{table}{
		Quantitative results.
		``Min.'' denotes the average value at the minima.
		For the saddle point values (``Sadd.''), the maximum value of each metric along the local MEPs is computed and the results are averaged.
		The ``epoch'' is measured at the point where the loss falls below the saddle point loss for the first time.
		It is noted in \textbf{bold} if it belongs to the third part of training with learning rate $\gamma=10^{-3}$.
		Basic CNNs and ResNets are trained for 136 epochs, DenseNets for 266 epochs.
		The ``factor'' is the ratio between average saddle point and minima loss.
		The standard deviations of all values at the minima and saddle are smaller than 10\% when averaged over the instances or over the mini-batches.
	}
	\centering
	\begin{small}
		\begin{sc}
			\begin{tabular}{ll|rrrr|rr|rrr}
\toprule
    &        & \multicolumn{4}{l}{Train energy} & \multicolumn{2}{l}{Test energy} & \multicolumn{3}{l}{Test error rate [\%]} \\
    &        &         Min. &   Sadd. & Factor &         Epoch &        Min. & Sadd. &                 Min. & Sadd. & $\Delta$ \\
dataset & architecture &              &         &        &               &             &       &                      &       &          \\
\midrule
C10+ & CNN-12 &  0.5428 &  0.6381 &  1.2 &  78 &  0.63 &  0.69 &  21.4 &  23.6 &  2.2 \\
    & CNN-24 &  0.4403 &  0.5390 &  1.2 &  84 &  0.59 &  0.64 &  19.8 &  21.8 &  2.0 \\
    & CNN-36 &  0.3982 &  0.4814 &  1.2 &  91 &  0.57 &  0.61 &  19.0 &  20.8 &  1.8 \\
    & CNN-48 &  0.3750 &  0.4331 &  1.2 &  \textbf{103} &  0.56 &  0.59 &  18.6 &  19.9 &  1.2 \\
    & CNN-96 &  0.3324 &  0.3900 &  1.2 &  \textbf{103} &  0.55 &  0.57 &  17.9 &  19.3 &  1.4 \\
    & CNN-48x2 &  0.1402 &  0.1564 &  1.1 &  \textbf{136} &  0.45 &  0.46 &  13.4 &  14.2 &  0.8 \\
    & CNN-48x3 &  0.0918 &  0.1164 &  1.3 &  \textbf{110} &  0.46 &  0.50 &  13.4 &  15.2 &  1.7 \\
    & ResNet-8 &  0.2045 &  0.2086 &  1.0 &  \textbf{136} &  0.37 &  0.38 &  12.0 &  12.5 &  0.4 \\
    & ResNet-20 &  0.0162 &  0.0324 &  2.0 &  \textbf{104} &  0.36 &  0.38 &  8.5 &  8.9 &  0.5 \\
    & ResNet-32 &  0.0057 &  0.0097 &  1.7 &  \textbf{107} &  0.36 &  0.37 &  7.5 &  7.8 &  0.3 \\
    & ResNet-44 &  0.0031 &  0.0060 &  1.9 &  \textbf{122} &  0.36 &  0.36 &  7.1 &  7.4 &  0.3 \\
    & ResNet-56 &  0.0022 &  0.0141 &  6.5 &  85 &  0.35 &  0.36 &  6.9 &  7.4 &  0.5 \\
    & DenseNet-40-12 &  0.0008 &  0.0046 &  5.9 &  \textbf{205} &  0.25 &  0.25 &  5.6 &  6.0 &  0.3 \\
    & DenseNet-100-12-BC &  0.0005 &  0.0026 &  4.8 &  \textbf{205} &  0.21 &  0.22 &  4.9 &  5.1 &  0.2 \\
C100+ & CNN-12 &  1.6167 &  1.9029 &  1.2 &  69 &  2.00 &  2.11 &  51.0 &  54.2 &  3.2 \\
    & CNN-24 &  1.3854 &  1.6930 &  1.2 &  70 &  1.90 &  1.99 &  48.2 &  51.5 &  3.3 \\
    & CNN-36 &  1.2670 &  1.5801 &  1.2 &  71 &  1.87 &  1.95 &  47.2 &  50.2 &  3.0 \\
    & CNN-48 &  1.1977 &  1.5002 &  1.3 &  73 &  1.87 &  1.94 &  46.6 &  49.5 &  2.9 \\
    & CNN-96 &  1.0549 &  1.3304 &  1.3 &  82 &  1.85 &  1.91 &  45.6 &  48.4 &  2.8 \\
    & CNN-48x2 &  0.6579 &  0.8372 &  1.3 &  91 &  1.71 &  1.73 &  41.0 &  42.6 &  1.6 \\
    & CNN-48x3 &  0.4393 &  0.6124 &  1.4 &  91 &  1.88 &  1.89 &  43.7 &  45.5 &  1.8 \\
    & ResNet-8 &  1.0894 &  1.1547 &  1.1 &  \textbf{103} &  1.46 &  1.49 &  39.6 &  40.6 &  1.0 \\
    & ResNet-20 &  0.3528 &  0.5442 &  1.5 &  79 &  1.42 &  1.48 &  33.3 &  34.7 &  1.4 \\
    & ResNet-32 &  0.1422 &  0.3576 &  2.5 &  77 &  1.53 &  1.57 &  31.5 &  33.7 &  2.2 \\
    & ResNet-44 &  0.0753 &  0.2117 &  2.8 &  85 &  1.60 &  1.62 &  30.8 &  32.5 &  1.7 \\
    & ResNet-56 &  0.0428 &  0.2978 &  7.0 &  71 &  1.64 &  1.66 &  30.3 &  32.4 &  2.0 \\
    & DenseNet-40-12 &  0.0101 &  0.0808 &  8.0 &  166 &  1.30 &  1.31 &  26.3 &  27.7 &  1.4 \\
    & DenseNet-100-12-BC &  0.0050 &  0.0223 &  4.4 &  \textbf{205} &  1.12 &  1.13 &  23.7 &  24.6 &  0.8 \\
\bottomrule
\end{tabular}

		\end{sc}
	\end{small}
	\label{tab:summary}
\end{table*}


\end{document}